\documentclass[pdflatex,sn-mathphys-num]{sn-jnl}

\usepackage{graphicx}%
\usepackage{multirow}%
\usepackage{amsmath,amssymb,amsfonts}%
\usepackage{amsthm}%
\usepackage{mathrsfs}%
\usepackage[title]{appendix}%
\usepackage{xcolor}%
\usepackage{textcomp}%
\usepackage{manyfoot}%
\usepackage{booktabs}%
\usepackage{algorithm}%
\usepackage{algorithmicx}%
\usepackage{algpseudocode}%
\usepackage{listings}%

\theoremstyle{thmstyleone}%
\theoremstyle{thmstyletwo}%

\theoremstyle{thmstylethree}%

\begin{document}

\title[Article Title]{Who Are You Behind the Screen? 

Implicit MBTI and Gender Detection Using Artificial Intelligence}


\author*[1]{\fnm{Kourosh} \sur{Shahnazari}}\email{kourosh@null.net}
\equalcont{These authors contributed equally to this work.}

\author[1]{\fnm{Seyed Moein} \sur{Ayyoubzadeh}}\email{smoein.ayyoubzadeh16@sharif.edu}
\equalcont{These authors contributed equally to this work.}

\affil*[1]{\orgdiv{Computer Engineering Department}, \orgname{Sharif University of Technology}}


\abstract{In personalized technology and psychological research, precisely detecting demographic features and personality traits from digital interactions becomes ever more important. This work investigates implicit categorization, inferring personality and gender variables directly from linguistic patterns in Telegram conversation data, while conventional personality prediction techniques mostly depend on explicitly self-reported labels. We refine a Transformer-based language model (RoBERTa) to capture complex linguistic cues indicative of personality traits and gender differences using a dataset comprising 138,866 messages from 1,602 users annotated with MBTI types and 195,016 messages from 2,598 users annotated with gender. Confidence levels help to greatly raise model accuracy to 86.16\%, hence proving RoBERTa's capacity to consistently identify implicit personality types from conversational text data. Our results highlight the usefulness of Transformer topologies for implicit personality and gender classification, hence stressing their efficiency and stressing important trade-offs between accuracy and coverage in realistic conversational environments. With regard to gender classification, the model obtained an accuracy of 74.4\%, therefore capturing gender-specific language patterns. Personality dimension analysis showed that people with introverted and intuitive preferences are especially more active in text-based interactions. This study emphasizes practical issues in balancing accuracy and data coverage as Transformer-based models show their efficiency in implicit personality and gender prediction tasks from conversational texts.}




\maketitle

\section{Introduction}

\subsection{The Myers-Briggs Type Indicator (MBTI)}

The Myers-Briggs Type Indicator (MBTI) is a widely utilized psychometric instrument in psychological research, therapy, corporate management, and educational settings for categorizing personality types based on individual preferences. The MBTI, further developed and systematized by Katharine Cook Briggs and her daughter Isabel Briggs Myers in the mid-20th century, stemmed from the psychological types proposed by Carl Gustav Jung in 1921. Analyzing natural tendencies across several psychological dimensions enables the characterization of human behavior, cognition, and social interactions based on foundational theoretical frameworks.\cite{myers1980gifts}

The MBTI evaluates personal inclinations across four opposing dimensions:
\begin{itemize} 
\item \textbf{Extraversion (E) - Introversion (I)} This dimension distinguishes an individual's focus on the exterior environment from their internal domain. Extraverted individuals typically succeed in social interactions, exhibiting active engagement with others and their environment. Introverted individuals gain energy and comfort from reflection, solitary activities, and reflective meditation, often requiring serene environments for productivity and personal growth.

\item \textbf{Sensing (S) – Intuition (N)} This dimension contrasts an individual's preference for tangible, empirical, and observable information (Sensing) with abstract, conceptual, and theoretical insights (Intuition). Individuals who prefer sensation rely heavily on direct sensory experiences and practical knowledge. Conversely, persons who choose intuition typically engage with concepts, patterns, and potential situations, often prioritizing innovation and theoretical frameworks over current realities.

\item \textbf{Thinking (T) – Feeling (F)}: This component reflects how individuals participate in decision-making processes. Individuals inclined towards Thinking prioritize logical, objective analysis and consistent principles, emphasizing rationality and impartiality. Individuals who select Feeling are guided by empathy, personal values, and the quest for emotional balance, emphasizing subjective experiences and social relationships.

\item \textbf{Judging (J) - Perceiving (P)}: This dimension indicates individuals' perspectives on structure, organization, and flexibility in their daily lives. Individuals who prioritize Judging typically favor structured, regulated, and organized environments, often exhibiting decisiveness and a desire for closure. In contrast, individuals who prefer Perceiving demonstrate flexibility, spontaneity, and receptiveness, often thriving in uncertainty and improvisation.

\end{itemize}

The synthesis of these four opposing attributes yields sixteen distinct personality types, represented by combinations such as INTJ, ESFP, INFP, or ESTJ. Each kind represents distinct behavioral patterns, cognitive styles, interpersonal dynamics, and professional or academic inclinations \cite{myers1980gifts}. Thus, accurately identifying and understanding an individual's MBTI type provides significant insights into their cognitive processes, preferences, and interpersonal dynamics, substantially improving psychological evaluation, career guidance, team leadership, and educational practices.

\subsection{Transformer-Based Language Models}

In the last ten years, substantial advancements have occurred in Natural Language Processing (NLP), primarily due to the emergence of Transformer-based neural network architectures. The Transformer, introduced by Vaswani et al. in 2017 \cite{vaswani2017attention}, has revolutionized language modeling by replacing traditional recurrent neural networks (RNNs) and convolutional neural networks (CNNs) with a framework exclusively based on self-attention mechanisms. This self-attention allows the model to dynamically evaluate the importance of different words inside a sentence or context, enhancing the modeling of semantic relationships and context-dependent linguistic features.

Three notable Transformer-based models have set new standards for several NLP tasks, including text classification, sentiment analysis, and language generation:

\begin{itemize}
\item \textbf{BERT (Bidirectional Encoder Representations from Transformers)}: Introduced by Devlin et al. in 2018 \cite{devlin2018bert}, BERT profoundly transformed NLP approaches by capturing contextual information bidirectionally. Through the utilization of masked language modeling and next-sentence prediction tasks during pre-training on large datasets, BERT attains an enhanced comprehension of context, semantics, and syntactic subtleties, leading to its widespread adoption in many NLP tasks, including question answering and text summarization.

\item \textbf{RoBERTa (Robustly Optimized BERT Approach)}: RoBERTa, developed by Liu et al. in 2019 \cite{liu2019roberta}, builds upon BERT by incorporating optimizations such as dynamic masking during training, an expanded training dataset, and the elimination of the next-sentence prediction target. These improvements enable RoBERTa to apprehend nuanced linguistic subtleties and contextual variability, consistently surpassing BERT and other baseline models across many NLP benchmarks.

\item \textbf{GPT-2 (Generative Pre-trained Transformer 2)}: GPT-2, introduced by Radford et al. \cite{radford2019language}, is fundamentally distinct from BERT and RoBERTa due to its unidirectional architecture, which is primarily focused on generating coherent and contextually pertinent text. GPT-2 has unparalleled capabilities in language generation, encompassing text completion and creative writing endeavors. Notwithstanding its generative emphasis, GPT-2 may be adeptly fine-tuned for discriminative classification tasks, illustrating the versatility of Transformer-based models.

\end{itemize}

The efficacy of Transformer models in natural language processing has significant ramifications, especially in utilizing linguistic analysis for personality and demographic profiling. By precisely capturing intricate linguistic patterns, these models offer a potential toolkit for assessing textual data to deduce psychological traits.

\subsection{Implicit Data Collection and Its Significance}

The significant proliferation of digital communication platforms such as social media, forums, and messaging applications has generated vast repositories of textual content created either overtly or implicitly by users. Analyzing this passive data stream has profound implications for psychological research, as it enables the covert extraction of personal characteristics, such as personality types and gender, without explicit self-reporting.

This method of implicit data collection presents numerous advantages:

\begin{itemize}
\item \textbf{Enhanced User Profiling}: Linguistic analysis improves user profiles and provides personality insights that improve service personalizing including customized learning platforms and targeted marketing strategies \cite{matz2017psychological}.

\item \textbf{Mental Health Monitoring}: Automated language analysis of user interactions helps to early detect and observe psychological disorders including depression or anxiety, so providing important opportunities for quick intervention and support \cite{de2013predicting}.

\item \textbf{Ethical and Privacy Considerations}: 
Although implicit data collecting has great possibilities, it also raises important ethical questions about user permission, privacy protection, and data governance. Thus, careful thought and open conversation on data use are absolutely crucial \cite{hinds2020contextual}.\cite{matz2017psychological}.

\end{itemize}

\subsection{Research Objectives}

Situated at the junction of computational linguistics and psychological assessment, this study has as its main goals:

\begin{itemize}
\item Evaluating the effectiveness of Transformer-based language models (BERT, RoBERTa, and GPT-2) for accurately classifying MBTI personality types and gender using textual data.
\item Investigating and identifying specific linguistic features and patterns that significantly contribute to accurate personality and gender classification.
\item Discussing the broader societal, ethical, and practical implications of employing sophisticated NLP models for implicit psychological profiling and passive demographic identification.
\end{itemize}

Intending to greatly advance both NLP and psychological science and so foster an interdisciplinary dialogue vital for responsible innovation and practical application in psychological profiling and personalized technology, this research analyzes extensive datasets including 138,666 messages from 1,602 users for MBTI classification and 195,016 messages from 2,598 users for gender classification.

\section{Related Work}

\subsection{Personality Prediction and MBTI Classification in Text}

Automated personality classification is a crucial subject in computational linguistics and applied natural language processing (NLP) for the psychological and social sciences. The Myers-Briggs Type Indicator (MBTI) is a prevalent psychological framework that classifies individuals into 16 unique personality types, derived from the intersections of four dichotomies: Introversion (I) vs. Extraversion (E), Sensing (S) vs. Intuition (N), Thinking (T) vs. Feeling (F), and Judging (J) vs. Perceiving (P). This framework renders MBTI a commonly utilized instrument for linguistic and behavioral analysis \citep{ashraf2024enhancing, li2024mbtibench}. Recent improvements in transformer designs have prompted studies to focus on utilizing deep learning models like BERT, RoBERTa, ALBERT, and DistilBERT for personality detection in textual data.

\subsection{Transformer-Based MBTI Classification}

Utilizing Transformer-based models for MBTI categorization has demonstrated significant advancements compared to conventional machine learning and lexicon-based approaches \citep{vasquez2021transformer}. Previous models predominantly utilized feature-based methodologies, including TF-IDF vectorization, LIWC dictionaries, and manually generated linguistic indicators. These approaches possessed intrinsic limitations, as they did not adequately capture contextual representations or semantic links within textual data.

Recent improvements indicate that pre-trained models, especially BERT-based architectures, exhibit superior performance in retrieving personality-related linguistic cues. Vásquez and Ochoa Luna \citep{vasquez2021transformer} investigated Transformer-based MBTI categorization utilizing the Kaggle dataset, attaining exemplary results with 88.63\% accuracy and 88.97\% F1-score. Their research established that contextual embeddings surpass conventional word-vector models, emphasizing BERT's capacity to discern subtle personality-related expressions across MBTI categories.

Research by does Santos and Paraboni \citep{dosSantos2022myers} shown that fine-tuning BERT for MBTI recognition on social media datasets markedly surpassed traditional classifiers, attaining strong generalization across several test settings. Kadambi \citep{kadambi2021exploring} validated these findings in their research on Twitter users with self-identified MBTI types, demonstrating that user profiles, status updates, and liked tweets uniquely enhance personality categorization accuracy.

\subsection{Limitations of Explicit MBTI Classification}

Most MBTI classification research employ datasets in which individuals explicitly self-report their personality type, resulting in intrinsic biases and concerns over data validity. Tareaf \citep{tareaf2022mbti} and Julianda and Maharani \citep{julianda2023personality} examined personality classification models developed using Reddit data, noting considerable class imbalances in MBTI distributions—certain personality types (e.g., ISTJ, ENFP) were markedly more prevalent than others, resulting in overfitting on majority-class samples.

To resolve this issue, Li et al. \citep{li2024mbtibench} developed MBTIBench, a meticulously maintained dataset employing soft-labeling techniques to mitigate inaccuracies arising from self-reported misclassifications. Their research revealed that approximately 30\% of data samples had erroneous user-assigned MBTI types, highlighting a significant issue with the dependability of self-reported personality datasets.

Notwithstanding enhancements from dataset curation methodologies like MBTIBench, this research continues to depend on explicit personality labels, which fail to translate effectively to implicit personality classification tasks in real-world scenarios, such as chat-based conversations.
\subsection{Implicit Personality Prediction in Conversational Text}

Unlike prior MBTI classification efforts, our work focuses on implicit personality prediction, where personality types are inferred without explicit self-reported labels. This presents multiple challenges:

\begin{itemize}
    \item \textbf{Absence of Direct Labels}: In contrast to Reddit and Twitter datasets, which feature user-provided MBTI annotations, Telegram data does not contain clear MBTI type disclosures. Consequently, personality assessment must deduce characteristics solely from linguistic styles, conversational involvement, and behavioral tendencies.
    
    \item \textbf{Noise and Informal Linguistic Structures}: Transforming natural conversation into effective model input involves substantial preprocessing, as Telegram messages contain slang, abbreviations, emojis, and multilingual text \citep{vonitsanos2023gender, dosSantos2022myers}.
    
    \item \textbf{Overlapping Personality and Gender Markers}: Several studies highlight correlations between MBTI dimensions and gender-linked linguistic traits, requiring careful disentangling of these features to avoid overfitting \citep{vonitsanos2023gender}.
\end{itemize}

\subsection{Transformer Fine-Tuning for Conversational Data}

Recent research highlights the need for robust fine-tuning procedures when applying Transformers to noisy data sources:

\begin{itemize}
    \item Arya et al. \citep{arya2023prediction} conducted comparative analyses on BERT, RoBERTa, DistilBERT, and ALBERT for personality classification, demonstrating that BERT-based models consistently produce the highest accuracy across social media text.
    
    \item Applications of hybrid filtering frameworks, including dynamic tokenization pipelines and structured pre-processing for informal digital conversations, are crucial for adapting models to real-world messaging platforms \citep{julianda2023personality}.
    
    \item Evaluation metrics must handle class imbalances dynamically. Prior studies apply macro-F1 and weighted precision-recall scores to counteract dataset bias. Methods such as oversampling, class reweighting, and contrastive learning have been explored as potential solutions \citep{ashraf2024enhancing, li2024mbtibench}.
\end{itemize}

\subsection*{The Need for Implicit Personality Modeling}

The prevalence of explicit self-reported MBTI classifications in previous studies considerably restricts the application of these models in authentic conversational contexts. Our research addresses this issue by (1) eliminating self-labeled training data, (2) enhancing tokenization and preprocessing methods for chat-based corpora, and (3) implementing Transformer-based fine-tuning techniques for implicit personality detection in Telegram conversations. This research transcends prior MBTI classification frameworks, facilitating the advancement of more resilient, real-time personality inference in text-based digital communication.

\vspace{0.386cm}
\section{Methodology}
\section*{Transformer Models: Mechanism and Evolution}

The emergence of \textbf{Transformer} models transformed natural language processing (NLP) by implementing a very efficient architecture founded on the self-attention mechanism. In contrast to conventional \textit{Recurrent Neural Networks (RNNs)} and \textit{Long Short-Term Memory (LSTM) networks}, Transformers eradicate sequential constraints by utilizing \textit{parallelized training} throughout whole input sequences. The basic design proposed by Vaswani et al. \citep{vaswani2017attention} employs a multi-layered self-attention mechanism that adaptively modifies weight distributions to encapsulate contextual word dependencies.

\subsection{Self-Attention and Multi-Head Attention}
In Transformer models, each input token contributes to the overall representation based on its \textit{attention scores} across all other tokens in a sequence. The attention score is computed using the following key operations:

\begin{equation}
    \text{Attention}(Q, K, V) = \text{softmax} \left( \frac{QK^T}{\sqrt{d_k}} \right) V
\end{equation}

where \( Q \) (Query), \( K \) (Key), and \( V \) (Value) matrices represent different transformations of input representations, and \( d_k \) is the scaling factor. \textbf{Multi-head attention} extends this mechanism by maintaining multiple attention weight matrices, allowing the model to capture diverse linguistic relationships.

\subsection{Pretraining Strategies: Masked Language Modeling and Next Sentence Prediction}
Most modern Transformer models rely on \textit{unsupervised pretraining techniques}, enabling them to learn extensive general-purpose linguistic features before fine-tuning on specific tasks. Notable pretrained architectures include:

\begin{itemize}
    \item \textbf{BERT (Bidirectional Encoder Representations from Transformers)} \citep{devlin2018bert}: Introduces \textbf{masked language modeling (MLM)}, where random tokens in input text are masked, forcing the model to predict masked words using surrounding context.
    \item \textbf{RoBERTa (Robustly Optimized BERT Approach)} \citep{liu2019roberta}: Enhances BERT’s training regime by removing next-sentence prediction (NSP) and training on significantly larger datasets.
    \item \textbf{GPT (Generative Pretrained Transformer)} \citep{radford2019language}: Utilizes a causal autoregressive decoding mechanism, allowing real-time text generation. Unlike BERT, GPT processes input unidirectionally.
\end{itemize}

\subsection{Expanding Transformer Applications in NLP}
Transformers have significantly advanced several NLP applications, including:

\begin{itemize}
    \item \textbf{Text Classification}: Models like BERT, RoBERTa, and DistilBERT dominate various classification tasks, from \textit{sentiment analysis} to \textit{document categorization} \citep{liu2019roberta}.
    \item \textbf{Machine Translation}: The Transformer model became the backbone of leading \textit{neural machine translation (NMT) systems}, outperforming LSTMs in real-time sentence translation \citep{vaswani2017attention}.
    \item \textbf{Conversational AI}: Large-scale Transformer models, such as \textit{GPT-based architectures}, power chatbots, virtual assistants, and real-time dialogue agents \citep{radford2019language}.
    \item \textbf{Psychological and Social Insights}: Transformers contribute to personality detection, author profiling, and \textit{mental health prediction via linguistic cues} \citep{de2013predicting, matz2017psychological}.
\end{itemize}

\subsection{Why Transformers Are Ideal for Personality and Gender Classification}
Several Transformer properties make them particularly well-suited for personality classification from text:

\begin{enumerate}
    \item \textbf{Bidirectional Context Awareness}: Unlike RNNs and LSTMs, BERT-based architectures process entire text sequences bidirectionally, capturing \textit{long-range dependencies} in user discourse.
    \item \textbf{Self-Attention for Linguistic Variability}: Personality and gender classification tasks benefit from \textit{self-attention mechanisms}, allowing models to highlight influential grammatical and stylistic cues.
    \item \textbf{Scalability and Adaptability}: Transformers can efficiently fine-tune on personality-labeled datasets while adapting to domain-specific text, such as \textit{Telegram chat-based data} \citep{julianda2023personality, vasquez2021transformer}.
\end{enumerate}

\subsection{The Need for Domain-Specific Transformer Adaptation}
While prior research has demonstrated the efficacy of Transformers in psychological linguistics, most studies use data from structured or explicitly labeled user-generated content (e.g., \textit{Reddit or Twitter MBTI-labeled datasets} \citep{kadambi2021exploring, dosSantos2022myers}). However, \textbf{Telegram conversations introduce unique linguistic challenges}:

\begin{itemize}
    \item \textbf{Multi-Turn Conversations and Threading}  
    Unlike Reddit, where posts exist independently, Telegram chat contexts evolve dynamically. Standard Transformer \textit{tokenization strategies struggle} to segment multi-turn dialogues effectively.
   
    \item \textbf{Noise and Non-Standard Grammar}  
    Informal messages include abbreviated words, emojis, and conversation artifacts that require \textit{more sophisticated text filtering techniques} than standard document classification tasks.
   
    \item \textbf{Ethical and Privacy Challenges}  
    Personality classification from chats raises ethical concerns regarding user profiling and data privacy \citep{hinds2020contextual}. Responsible data anonymization, differential privacy techniques, and user consent protocols should inform future work.
\end{itemize}

\subsection*{Integrating Transformer Advances for Implicit Personality Detection}
The Transformer revolution has enabled substantial progress in NLP-based personality classification; nonetheless, current research predominantly centers on explicit personality disclosures, such as self-reported MBTI classifications. Our research seeks to enhance the discipline by:

\begin{itemize} 
\item Utilizing the contextual representations of Transformers for chat-based implicit personality inference.
\item Enhancing preprocessing/tokenization techniques to accommodate informal Telegram communication. 
\item Guaranteeing comprehensive model evaluation through multi-layered fine-tuning strategies. \end{itemize}

The integration of self-attention architectures, unsupervised personality modeling, and informal conversational adaptations will facilitate the development of more realistic psychological AI applications, connecting theoretical NLP advancements with practical behavioral insights in real-world dialogues.

\section{Overview of the Transformer Architecture}
The Transformer model follows an \textbf{encoder-decoder} structure:
\begin{enumerate}
    \item The \textbf{encoder} processes input sequences into contextual representations.
    \item The \textbf{decoder} generates output sequences by attending to encoder outputs and previous generated tokens.
\end{enumerate}
Both components consist of multiple stacked layers, incorporating:
\begin{itemize}
    \item \textbf{Multi-head self-attention mechanisms}
    \item \textbf{Feedforward networks}
    \item \textbf{Layer normalization and residual connections}
\end{itemize}

\section{Step-by-Step Breakdown of the Transformer Algorithm}
\subsection{Input Embedding and Positional Encoding}
Since Transformers \textbf{do not have recurrence}, they require a method to retain positional order in sequences. The model uses:
\begin{enumerate}
    \item Learned \textbf{word embeddings} to convert input tokens into dense vectors.
    \item \textbf{Positional encoding} to introduce order information, computed as follows:
    \begin{equation}
        PE_{\text{pos}, 2i} = \sin\left(\frac{\text{pos}}{10000^{2i/d_{\text{model}}}}\right)
    \end{equation}
    \begin{equation}
        PE_{\text{pos}, 2i+1} = \cos\left(\frac{\text{pos}}{10000^{2i/d_{\text{model}}}}\right)
    \end{equation}
    where $\text{pos}$ represents the position index and $d_{\text{model}}$ is the dimensionality of embeddings.
\end{enumerate}

\subsection{Encoder Mechanism}
Each encoder layer performs the following:

\subsubsection{Multi-Head Self-Attention}
Each token representation is transformed into three matrices:
\begin{equation}
    Q = X W_Q, \quad K = X W_K, \quad V = X W_V
\end{equation}
where:
\begin{itemize}
    \item \( Q \) (Query) determines what information to extract.
    \item \( K \) (Key) contains relevance scores against other tokens.
    \item \( V \) (Value) represents token information passed through attention weightings.
\end{itemize}

The \textbf{scaled dot-product attention} is computed as:
\begin{equation}
    \text{Attention}(Q, K, V) = \text{softmax} \left( \frac{QK^T}{\sqrt{d_k}} \right) V
\end{equation}
where the denominator \( \sqrt{d_k} \) prevents instability in gradient updates.

\subsubsection{Residual Connections and Layer Normalization}
To stabilize training, the model adds a \textbf{residual connection}:
\begin{equation}
    Z = \text{LayerNorm}(A + X)
\end{equation}
Following this, a \textbf{feedforward network} enhances non-linear transformations:
\begin{equation}
    O = \text{ReLU}(Z W_1 + b_1) W_2 + b_2
\end{equation}
The final encoder output is normalized:
\begin{equation}
    E_X = \text{LayerNorm}(O + Z)
\end{equation}

\subsection{Decoder Mechanism}
The decoder consists of similar layers to the encoder but adds:
\begin{enumerate}
    \item \textbf{Masked self-attention} to prevent information leakage from future tokens.
    \item \textbf{Cross-attention with encoder outputs} to integrate input sequence information.
\end{enumerate}

\subsubsection{Masked Attention}
To ensure autoregressive sequence generation, future tokens are masked:
\begin{equation}
    A = \text{MaskedMultiHeadAttention}(Q, K, V)
\end{equation}

\subsubsection{Encoder-Decoder Cross-Attention}
Each decoder token attends to encoder outputs:
\begin{equation}
    A' = \text{MultiHeadAttention}(Q', K', V')
\end{equation}
where:
\begin{itemize}
    \item \( Q' \) comes from the decoder.
    \item \( K' \) and \( V' \) come from the encoder outputs.
\end{itemize}

\subsection{Final Output Computation}
The decoder output is passed through a Softmax layer to compute word probabilities:
\begin{equation}
    P(y_t) = \text{Softmax}(E_Y W_o)
\end{equation}
The predicted sequence is obtained via:
\begin{equation}
    \hat{Y} = \arg\max P(y_t)
\end{equation}

\begin{algorithm}
\caption{Transformer: Encoder-Decoder Architecture}
\begin{algorithmic}[1]
\Require Input sequence $X = (x_1, x_2, ..., x_n)$, Target sequence $Y = (y_1, y_2, ..., y_m)$
\Ensure Transformed output sequence $\hat{Y}$

\State \textbf{Initialization:} Load parameters $\theta$ for Encoder and Decoder networks
\State \textbf{Compute input embeddings:} $E_X \gets \text{Embed}(X) + \text{PosEnc}(X)$
\State \textbf{Compute target embeddings:} $E_Y \gets \text{Embed}(Y) + \text{PosEnc}(Y)$

\For {\textbf{each encoder layer} $l = 1$ to $L$}
    \State $Q, K, V \gets E_X W_Q^l, E_X W_K^l, E_X W_V^l$ \Comment{Linear projections}
    \State $A \gets \text{MultiHeadAttention}(Q, K, V)$ \Comment{Self-Attention}
    \State $Z \gets \text{LayerNorm}(A + E_X)$ \Comment{Residual connection and Layer Normalization}
    \State $O \gets \text{FeedForward}(Z)$ \Comment{Position-wise feedforward network}
    \State $E_X \gets \text{LayerNorm}(O + Z)$ \Comment{Final residual connection}
\EndFor

\State Encoder output: $H \gets E_X$

\For {\textbf{each decoder layer} $l = 1$ to $L$}
    \State $Q, K, V \gets E_Y W_Q^l, E_Y W_K^l, E_Y W_V^l$
    \State $A \gets \text{MaskedMultiHeadAttention}(Q, K, V)$ \Comment{Masked to prevent seeing future tokens}
    \State $Z \gets \text{LayerNorm}(A + E_Y)$
    \State $Q' \gets Z W_{Q'}^l, K' \gets H W_{K'}^l, V' \gets H W_{V'}^l$
    \State $A' \gets \text{MultiHeadAttention}(Q', K', V')$ \Comment{Cross-Attention with encoder output}
    \State $Z' \gets \text{LayerNorm}(A' + Z)$
    \State $O \gets \text{FeedForward}(Z')$
    \State $E_Y \gets \text{LayerNorm}(O + Z')$
\EndFor

\State Decoder output: $\hat{Y} \gets \text{Softmax}(E_Y W_{o})$ \Comment{Final token probabilities}

\State \Return $\hat{Y}$
\end{algorithmic}
\end{algorithm}

\subsection{Dataset and Preprocessing}
Our study utilizes an anonymized dataset of messages extracted from a Telegram social channel. Users have self-reported their MBTI personality type and gender. Unlike structured datasets, this dataset consists of naturally occurring conversations, allowing us to explore implicit personality and gender classification through linguistic features.

\begin{figure}[h]
    \centering
    \includegraphics[width=0.9\linewidth]{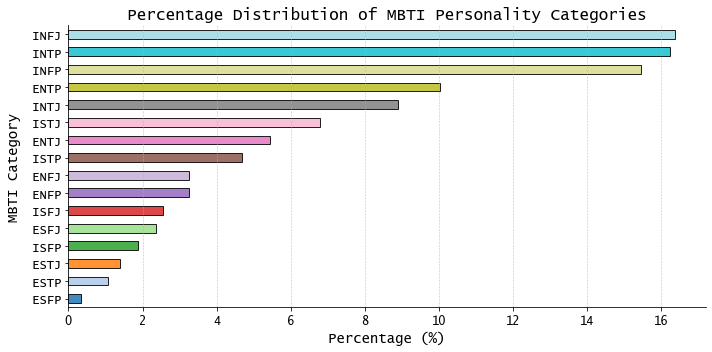}
    \caption{Distribution of messages across MBTI types and gender categories in the dataset.}
    \label{fig:dataset_distribution}
\end{figure}

\subsection*{Dataset Composition}
The dataset is divided into two primary tasks:
\begin{itemize}
    \item \textbf{MBTI Classification}: 138,866 messages from 1,602 users with annotated MBTI labels.
    \item \textbf{Gender Classification}: 195,016 messages from 2,598 users with annotated gender labels.
\end{itemize}
\noindent
Each message contains:
\begin{enumerate}

    \item The raw message text.
    \item An associated MBTI type (if available).
    \item An associated gender label (if available).
\end{enumerate}
Some messages include both MBTI and gender labels, while others contain only one of these attributes.
\subsubsection*{MBTI Feature Distribution}

We show a representation of the distribution of MBTI personality traits in our dataset, focusing on the four key dichotomies: Extroversion/Introversion, Sensing/Intuition, Thinking/Feeling, and Judging/Perceiving. Comprehending these distributions is essential.

\begin{figure}[h]
    \centering
    \includegraphics[width=0.8\linewidth]{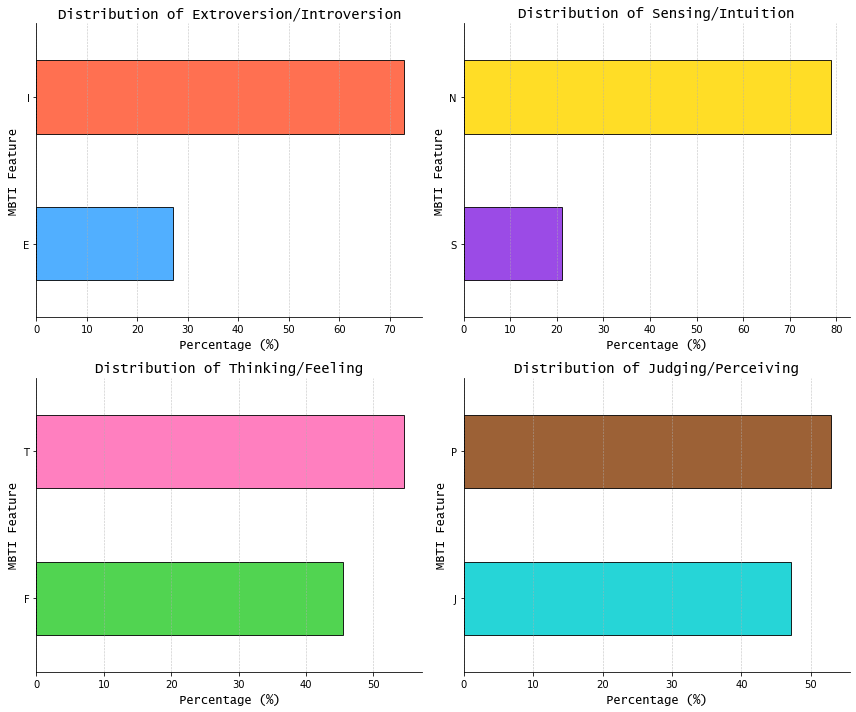}
    \caption{Distribution of MBTI personality dichotomies within the dataset. Each subplot represents the proportion of messages associated with a specific MBTI feature.}
    \label{fig:mbti_distribution}
\end{figure}

The distribution of MBTI personality traits in our sample, illustrated in Figure~\ref{fig:mbti_distribution}, demonstrates significant variability across the four dichotomous dimensions.

These discrepancies may be shaped by the intrinsic preferences of persons possessing particular personality traits regarding computer-mediated communication (CMC), such as messaging platforms.

\subsubsection{Introversion (I) Preference}

Individuals exhibiting Introverted (I) inclinations frequently prefer textual communication to in-person contact. This desire enables individuals to regulate social interactions at their own tempo, along with their propensity for introspection and deliberate self-presentation. Studies suggest that introverts tend to favor online communication approaches, as these platforms allow for contemplative processing prior to responding \cite{harrington2010mbti}. As a result, introverts may exhibit more activity on messaging platforms, resulting in a greater representation of Introversion (I) within the dataset.

\subsubsection{Intuition (N) Prevalence}

Individuals with an intuitive preference are drawn to abstract concepts and future possibilities, often engaging in discussions that explore theoretical ideas and patterns. Digital communication platforms Online messaging applications are well-suited for investigations, as they enable extensive discussions at any time, allowing individuals to articulate their thoughts without immediate concerns regarding the practical implications of their statements. The discussion may result in a higher representation of Intuitive (N) users within our sample.

\subsubsection{Thinking (T) Dominance}

The higher prevalence of Thinking (T) relative to Feeling (F) in the sample can be attributed to the nature of online communication, which often emphasizes information sharing and discussion over emotional expression. Individuals who prioritize logic and objectivity may find messaging systems advantageous for participating in analytical and problem-solving discussions, thereby improving their engagement and representation.

\subsubsection{Perceiving (P) Notability}

Perceiving (P) individuals are characterized by adaptability and spontaneity, traits that align well with the dynamic and flexible nature of online messaging. The lack of rigid structure in CMC allows Perceivers (P) to navigate conversations fluidly, aligning with their preference for open-endedness and improvisation. This compatibility may contribute to the notable presence of Perceiving (P) users in the dataset.

\begin{figure}[h]
    \centering
    \includegraphics[width=0.8\linewidth]{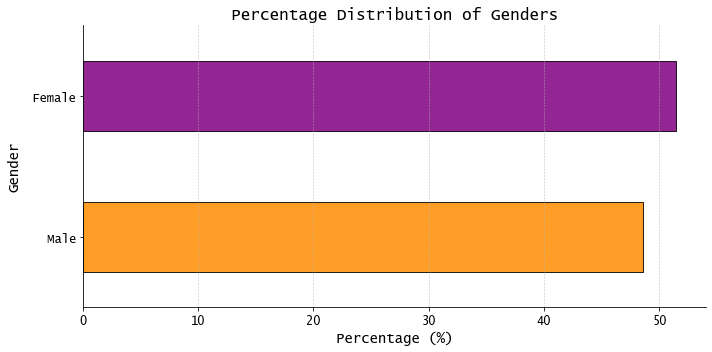}
    \caption{Percentage distribution of genders in the dataset. The chart represents the proportion of messages attributed to male and female users.}
    \label{fig:gender_distribution}
\end{figure}

\noindent
The dataset's distribution of MBTI personality types, illustrated in Figure~\ref{fig:dataset_distribution}, shows significant differences compared to general population distributions. INFJ is identified as the most common type, with INTP and INFP following closely, whereas ESFP exhibits the lowest representation. This pattern reveals distinct differences in communication styles driven by personality, self-selection biases, and interaction preferences in online messaging forums.

INFJs, representing roughly 16.37\% of the dataset, exhibit a pronounced preference for text-based communication, attributed to their introspective characteristics and inclination towards profound, meaningful interactions. Their inclination towards structured and emotionally engaging discussions corresponds with the nature of online forums, enabling them to articulate their insights and values thoughtfully, free from immediate social pressures.

INTPs account for 16.22\%, indicative of their analytical and intellectually curious characteristics. INTPs are drawn to platforms that provide intellectual stimulation, excelling in settings that allow for the analysis of complex ideas and participation in theoretical discussions. However, their generally selective participation style may result in a reduced volume of interactions compared to INFJs, who tend to engage more frequently due to their relational approach.

INFPs comprise 15.44\% of the dataset, reflecting a notable presence attributed to their inclination towards self-expression and the exploration of abstract concepts. Their introspective and emotionally expressive communication style effectively engages online forums through detailed and reflective posts. This corresponds with their established inclination to utilize written communication as a means of examining personal insights and emotions.

ENTP, at 10.02\%, and INTJ, at 8.89\%, exemplify personality types characterized by a focus on intellectual engagement and abstract reasoning. ENTPs exhibit a strong inclination towards debate and the exploration of ideas, often taking the initiative or engaging actively in dynamic discussions. INTJs favor conceptual and strategic discussions, utilizing forums as platforms for idea exchange, free from the distractions of in-person interactions.

ISTJ (6.79\%) and ENTJ (5.43\%) types exhibit moderate levels of engagement. The communication style of ISTJs is characterized by practicality and structure, which may restrict their engagement in abstract or emotionally charged discussions, as they tend to prioritize factual and clearly defined exchanges. ENTJs typically adopt a leadership-oriented approach that emphasizes strategic contributions over frequent participation, reflecting their goal-directed and succinct communication style.

ESFJ (2.37\%) and ISFJ (2.55\%) demonstrate lower representation, consistent with their inclination towards direct, socially structured interactions. Their focus on face-to-face and relational communication may hinder substantial involvement in online text-based discussions, which lack interpersonal cues.

The least represented types—ESTJ (1.40\%), ESTP (1.08\%), and ESFP (0.34\%)—exhibit reduced activity levels, attributed to their pronounced inclination towards immediate, pragmatic, and socially interactive environments. These personality types emphasize direct interaction, active engagement, and sensory experiences, which diminishes their tendency to participate in written and reflective online discussions.

The MBTI type distribution in our dataset highlights the impact of personality traits on digital communication behaviors. Types defined by introspection, abstraction, and a preference for textual engagement (INFJ, INTP, INFP) are significantly overrepresented, whereas action-oriented, socially driven types (ESFP, ESTP, ESTJ) are notably underrepresented. The findings underscore significant personality-driven variations that affect online participation.

\subsubsection*{Gender Distribution}

The dataset's gender distribution, illustrated in Figure~\ref{fig:gender_distribution}, presents the ratio of messages from male and female users. The dataset includes contributions from both genders, exhibiting no significant dominance of one category over the other. This balanced distribution allows the model to encounter a variety of linguistic patterns linked to different genders. 

Linguistic research indicates that gender differences in language use can be observed through variations in sentence structure, word choice, and conversational style. These distinctions may function as valuable attributes for the gender classification task. The identification of gender-specific linguistic patterns within the dataset allows Transformer-based models to effectively discern features that differentiate male and female communication styles.

The dataset offers a comprehensive representation of both genders; however, it is crucial to recognize that language use is shaped by various factors beyond gender, such as personality, cultural background, and social context. Consequently, although gender-based linguistic tendencies may appear in model predictions, they should not be regarded as definitive indicators of gender identity. The classification model learns probabilistic relationships between textual features and the assigned gender labels.

The insights derived from this distribution enhance our understanding of the influence of gender representation in textual data on classification performance. Future research may examine biases in gender prediction models and assess the extent to which specific linguistic features disproportionately influence classification outcomes.

\subsubsection{Preprocessing Pipeline}
Given Telegram messages' informal and unstructured nature, substantial preprocessing was performed to standardize text and reduce noise. Our preprocessing pipeline includes:
\begin{enumerate}
    \item \textbf{Removing Links}: All hyperlinks were removed to eliminate irrelevant content.
    \item \textbf{Filtering Short Messages}: Messages with fewer than 25 tokens were discarded to ensure meaningful linguistic representation.
    \item \textbf{Eliminating Explicit MBTI Mentions}: Any direct mention of MBTI types (e.g., "I am an ISTJ") was removed to prevent information leakage.
    \item \textbf{Tokenization}: We applied model-specific tokenization methods, such as WordPiece for BERT and Byte-Pair Encoding (BPE) for GPT-2.

\end{enumerate}
\noindent
These preprocessing steps ensure that only relevant linguistic features contribute to model training.

\subsection{Model Selection and Feature Extraction}
To classify MBTI types and gender, we employ Transformer-based models, extracting high-dimensional feature vectors from their embedding layers.

\subsubsection{Transformer Models for Feature Extraction}
We utilize the following pre-trained Transformer architectures:
\begin{itemize}
    \item \textbf{BERT (Bidirectional Encoder Representations from Transformers)}
    \item \textbf{RoBERTa (Robustly Optimized BERT Pretraining Approach)}
    \item \textbf{GPT-2 (Generative Pretrained Transformer 2)}
\end{itemize}

For each model, we extract feature vectors from the final embedding layer, capturing contextual linguistic representations. The extracted feature vectors serve as input for classification.

\subsubsection{Feature Vector Representation}
The extracted embeddings are processed with the following configuration:
\begin{itemize}
    \item \textbf{Hidden size}: 768
    \item \textbf{Number of hidden layers}: 12
    \item \textbf{Number of attention heads}: 12
    \item \textbf{Output vector dimensionality}: 16
\end{itemize}

These embeddings encapsulate complex linguistic structures and are used as input features for MBTI and gender classification.

\subsection{Fine-Tuning Strategy}
Each model undergoes fine-tuning using a classification head attached to the final embedding representation. The fine-tuning process follows these steps:
\begin{enumerate}
    \item \textbf{Tokenization}: Messages are tokenized using the model’s corresponding tokenizer.
    \item \textbf{Feature Extraction}: Hidden-state embeddings from the final layer are used as classification inputs.
    \item \textbf{Dimensionality Reduction}: The extracted 768-dimensional embeddings are reduced to a 16-dimensional representation.
    \item \textbf{Classification Head}: A fully connected layer with a softmax activation function predicts the target label.
\end{enumerate}

The loss function for fine-tuning is categorical cross-entropy, defined as:

\begin{equation}
    L = -\sum_{i=1}^{N} y_i \log(\hat{y}_i)
\end{equation}

where \( y_i \) is the true label and \( \hat{y}_i \) is the predicted probability.

\subsubsection{Training Configuration}
The training process is configured as follows:
\begin{itemize}
    \item \textbf{Batch size}: 64
    \item \textbf{Learning rate}: \( 2.86e^{-5} \), optimized using AdamW
    \item \textbf{Number of epochs}: 16
    \item \textbf{Max length (Tokens)}: 128
    \item \textbf{Graphics Processing Unit (GPU)}: Nvidia Tesla T4
\end{itemize}                                                                                                                                            

\subsection{Evaluation Metrics}
To assess classification performance, we utilize:
\begin{enumerate}
    \item \textbf{Accuracy}: Measures the proportion of correctly classified instances.
    \item \textbf{Precision, Recall, and F1-Score}: Evaluates class-specific performance, balancing false positives and false negatives.
    \item \textbf{Macro-F1 Score}: Accounts for class imbalance by averaging F1-scores across all MBTI types.
    \item \textbf{Confusion Matrix Analysis}: Identifies misclassification patterns and model biases.
\end{enumerate}

\section{Results}

In this section, we present the performance evaluation of our models on MBTI personality classification and gender classification tasks. We report accuracy, precision, recall, and F1-score as key performance metrics. Table~\ref{tab:mbti_results} and Table~\ref{tab:gender_results} summarize our findings.

\subsection{MBTI Personality Classification}

Table~\ref{tab:mbti_results} presents the performance of BERT, GPT-2, and RoBERTa on the MBTI personality classification task. Among these models, RoBERTa demonstrates the highest classification performance, achieving an accuracy of \textbf{49\%} and an F1-score of \textbf{50\%}. This suggests that RoBERTa effectively captures the linguistic nuances associated with different MBTI personality types, outperforming both BERT and GPT-2.

\begin{table}[h]
    \centering
    \begin{tabular}{lcccc}
        \hline
        \textbf{Model} & \textbf{Accuracy} &  \textbf{Precision} & \textbf{Recall} & \textbf{F1-Score}  \\
        \hline
        BERT & 40\% & 39\% & 40\% & 0.39 \\
        GPT-2 & 45\% & 45\% & 45\% & 0.44 \\
        RoBERTa & \textbf{49\%} & \textbf{51\%} & \textbf{49\%} & \textbf{0.50}  \\
        \hline
    \end{tabular}
    \caption{Performance comparison for MBTI personality classification.}
    \label{tab:mbti_results}
\end{table}

\begin{figure}[h]
    \centering
    \includegraphics[width=0.8\linewidth]{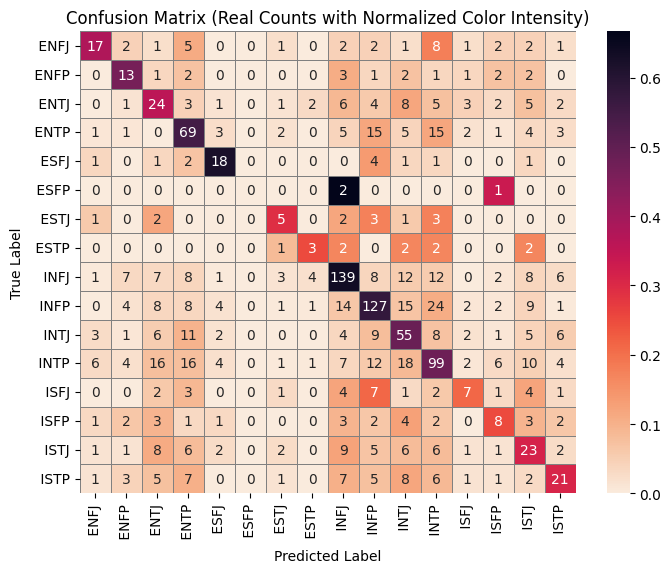}
    \caption{Confusion matrix for RoBERTa on MBTI personality classification.}
    \label{fig:conf_matrix}
\end{figure}

To further analyze the model's classification behavior, we visualize the confusion matrix for RoBERTa in Figure~\ref{fig:conf_matrix}. The confusion matrix provides insight into misclassifications across MBTI types, revealing that certain personality categories, particularly those with overlapping linguistic patterns, are more prone to misclassification.

\subsection{Effect of Confidence Thresholds on MBTI Classification}
In standard classification scenarios, models assign the class with the highest probability to each instance, irrespective of the confidence level of the prediction. In practical applications, particularly those related to implicit personality classification, the reliability of predictions is essential. In response to this issue, we implemented a confidence threshold parameter, categorizing only instances with predicted confidence surpassing a defined threshold.

Modifying the confidence threshold serves as an effective method for markedly improving prediction accuracy. RoBERTa effectively captures robust linguistic indicators that are strongly associated with specific MBTI personality types by selectively classifying instances with higher confidence scores. The results indicate that at a confidence threshold of 0.99, RoBERTa attained an accuracy of 86.16\%. This significant enhancement highlights the model's efficacy in recognizing and utilizing linguistic patterns that are indicative of specific personality traits.

Table~\ref{tab:confidence_threshold} presents the relationship between different confidence thresholds, classification accuracy, and the proportion of classified data. As shown, increasing the confidence threshold leads to a notable increase in classification accuracy. However, this comes at the cost of reduced data coverage. For example, at a threshold of 0.50, the model achieves 52.83\% accuracy while covering 85\% of the dataset, whereas at the highest threshold of 0.99, accuracy reaches 86.16\% but covers only 26\% of the dataset.

\begin{table}[h]
    \centering
    \caption{Effect of Confidence Thresholds on MBTI Classification}
    \label{tab:confidence_threshold}
    \begin{tabular}{c|c|c}
        \hline
        \textbf{Threshold} & \textbf{Accuracy (\%)} & \textbf{Data Coverage (\%)} \\
        \hline
        0.50 & 52.83 & 85 \\
        0.60 & 55.51 & 77 \\
        0.70 & 59.82 & 68 \\
        0.80 & 64.12 & 59 \\
        0.90 & 70.16 & 48 \\
        0.99 & 86.16 & 26 \\
        \hline
    \end{tabular}
\end{table}

The high accuracy observed at elevated thresholds has significant implications, particularly in domains where precision is critical, such as personalized psychological counseling, targeted marketing, and educational settings. Employing elevated confidence thresholds improves reliability in automated personality predictions, guaranteeing that the outcomes or insights generated by the model are precise and applicable.

Adopting higher confidence thresholds leads to a reduction in the number of classified data points, highlighting a trade-off between accuracy and data coverage. As seen in Table~\ref{tab:confidence_threshold}, at the highest threshold of 0.99, approximately 26\% of the total dataset remains classified. This suggests that although RoBERTa's predictions exhibit high accuracy at increased confidence levels, a significant amount of data remains unclassified.

Nevertheless, this limitation is offset by the immense value that such a highly accurate classification provides. The ability to predict an individual’s MBTI personality type with 86.16\% accuracy, solely based on their text messages, is a remarkable achievement in the field of computational psychology and AI-driven personality analysis. This level of precision suggests that even in real-world applications where only a fraction of messages can be confidently classified, the insights generated by the model remain highly reliable and valuable.

Such a high-confidence personality prediction system can be particularly useful in applications where accurate personality profiling is essential, such as in psychological assessments, AI-driven recruitment systems, and adaptive learning platforms that personalize educational content based on inferred personality traits. The ability to infer personality traits implicitly, without requiring users to take explicit psychological tests, not only enhances user experience but also opens the door to more seamless and non-intrusive applications in personalized AI systems.

Furthermore, our results demonstrate the substantial benefits of confidence-threshold tuning in transformer-based implicit personality classification tasks, emphasizing the methodological rigor and practical relevance of our approach. By dynamically adjusting the confidence threshold, practitioners can optimize their models for either broad coverage or high precision, depending on their specific application requirements.

A significant avenue for future research involves utilizing multiple messages per user instead of focusing on single messages, which may enhance both prediction accuracy and coverage through aggregation. Considering multiple conversational samples per individual allows models to address the coverage limitation seen at high confidence thresholds, thereby improving the practical applicability and robustness of implicit personality detection from conversational data. By leveraging larger conversational contexts, it is likely that we can push classification accuracy even further while maintaining broader coverage.

\subsection{MBTI Subtype Classification}

Given RoBERTa’s superior performance in MBTI classification, we further evaluate its ability to classify individual MBTI dimensions (E/I, S/N, T/F, J/P). Table~\ref{tab:subtype_results} presents the F1-scores for each dichotomy, showing that the model achieves competitive performance across all four personality dimensions.

\begin{table}[h]
    \centering
    \begin{tabular}{lcccc}
        \hline
        \textbf{MBTI Dichotomy} & \textbf{Accuracy} & \textbf{Precision} & \textbf{Recall} & \textbf{F1-Score} \\
        \hline
        Extraversion / Introversion (E/I) & 79\% & 77\% & 79\% & 0.77 \\
        Sensing / Intuition (S/N) & 81\% & 78\% & 81\% & 0.78 \\
        Thinking / Feeling (T/F) & 71\% & 71\% & 71\% & 0.70 \\
        Judging / Perceiving (J/P) & 71\% & 71\% & 71\% & 0.71 \\
        \hline
    \end{tabular}
    \caption{Performance of RoBERTa for MBTI subtype classification.}
    \label{tab:subtype_results}
\end{table}

\subsection{Gender Classification}

We also evaluated RoBERTa on the gender classification task. Table~\ref{tab:gender_results} summarizes the precision, recall, and F1-score for classifying gender from textual data. RoBERTa achieved an accuracy of \textbf{74.40\%}, demonstrating balanced performance across both classes.

\begin{table}[htbp] \centering \begin{tabular}{lccc} \hline \textbf{Gender} & \textbf{Precision} & \textbf{Recall} & \textbf{F1-Score} \\ \hline Female & 73\% & 77\% & 0.75 \\ Male & 76\% & 72\% & 0.74 \\ \hline Macro Avg. & 75\% & 74\% & 0.74 \\ Weighted Avg. & 75\% & 74\% & 0.74 \\ \hline \end{tabular} \caption{Classification performance of RoBERTa on gender prediction.} \label{tab:gender_results} \end{table}

\begin{figure}[h] \centering \includegraphics[width=0.63\linewidth]{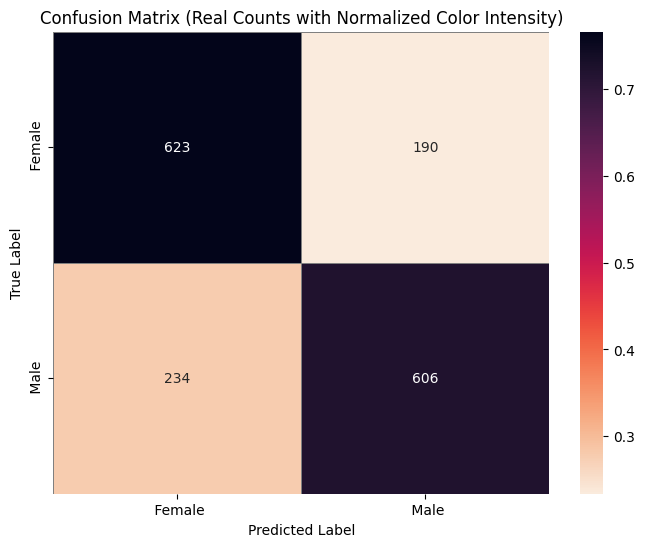} \caption{Confusion matrix for RoBERTa on gender classification.} \label{fig:gender_confusion_matrix} \end{figure}

The confusion matrix shown in Figure~\ref{fig:gender_confusion_matrix} further clarifies the classification behavior of RoBERTa. The model correctly identified 77\% of females and 72\% of males, with a relatively balanced distribution of misclassifications between both genders. This suggests RoBERTa effectively captures gender-specific linguistic patterns but indicates potential areas of overlap where communication styles between genders are not distinctly separable.

Overall, these results confirm RoBERTa’s capability in distinguishing gender-based linguistic nuances within text data, although further improvement in distinguishing subtle differences remains necessary.

\subsection{Analysis of Model Performance}

Because of its optimal training schedule—which incorporates dynamic masking, bigger pretraining datasets, and the deletion of the next-sentence prediction objective—which so improves generalization-RoBERTa shows better performance across all classification tasks. Unlike GPT-2, which emphasizes generative tasks, RoBERTa models contextual relationships necessary for categorization by means of bidirectional encoding.

The subtype classification task demonstrates superior performance relative to the exact MBTI type classification. This corresponds with the hierarchical structure of MBTI traits, wherein categorizing broad personality dimensions (e.g., E/I) is fundamentally more straightforward than forecasting a 16-class outcome. Furthermore, gender classification demonstrates superior accuracy compared to MBTI classification, likely attributable to more pronounced linguistic differences linked to gender.

\section{Conclusion}

This study examines the effectiveness of Transformer-based language models, particularly RoBERTa, in classifying MBTI personality types and gender through conversational textual data sourced from Telegram forums. The findings demonstrate the considerable potential of utilizing linguistic cues from informal conversational contexts for implicit personality and demographic profiling.

Our findings indicate that RoBERTa successfully identifies personality-specific linguistic features, outperforming conventional methods in MBTI personality classification. The implementation of confidence thresholds emphasized the model's capacity to enhance prediction accuracy while sacrificing coverage, illustrating a significant trade-off relevant in precision-sensitive contexts.

In gender classification, RoBERTa demonstrated proficiency by effectively distinguishing gender-specific linguistic patterns, exhibiting balanced performance across classes. The analysis of the confusion matrix revealed distinct linguistic differences between male and female users, though some overlap in language patterns indicates that further refinement may enhance model differentiation.

Our analysis of MBTI feature distributions in conversational text highlights the substantial impact of personality-driven communication preferences on user interactions in digital platforms. The prevalence of introverted and intuitive personality types highlights their suitability for online text-based communication, whereas the lesser presence of sensing, judging, and extroverted types indicates a preference for face-to-face, structured interactions rather than digital engagement.

Although RoBERTa has shown effectiveness, challenges persist in implicit personality classification within informal conversational data. Challenges including noisy and unstructured language, lack of explicit labeling, and ethical concerns related to user privacy and data utilization remain prevalent. To address these challenges, advanced preprocessing techniques, adaptive tokenization strategies, and ethical frameworks are necessary to ensure the responsible use of implicit psychological modeling.

Future research should conduct in-depth analyses of linguistic features associated with specific personality and gender dimensions, create advanced fine-tuning strategies specifically designed for conversational data, and examine ethical implications in greater detail. 

\section*{Declarations}
The authors declare that no ethical guidelines were violated, and no personally identifiable information was accessed or disclosed in conducting this research.

\subsection*{Ethical Considerations and Compliance}

All data utilized in this study were collected in full compliance with the Telegram platform's Terms of Service. Users’ privacy and confidentiality were strictly maintained through anonymization procedures.

\subsection*{Consent and Data Usage}

All participants implicitly consented to the use of their anonymized messages for analytical purposes through their agreement with Telegram's Terms of Service. No personally identifiable information was disclosed or analyzed during this study.

\subsection*{Conflict of Interest}

The authors declare no conflicts of interest regarding the publication of this research.

\subsection*{Ethical Approval}

This research adhered strictly to ethical guidelines regarding data handling and privacy.

\bigskip

\bibliography{sn-bibliography}


\begin{thebibliography}{18}
\ifx \bisbn   \undefined \def \bisbn  #1{ISBN #1}\fi
\ifx \binits  \undefined \def \binits#1{#1}\fi
\ifx \bauthor  \undefined \def \bauthor#1{#1}\fi
\ifx \batitle  \undefined \def \batitle#1{#1}\fi
\ifx \bjtitle  \undefined \def \bjtitle#1{#1}\fi
\ifx \bvolume  \undefined \def \bvolume#1{\textbf{#1}}\fi
\ifx \byear  \undefined \def \byear#1{#1}\fi
\ifx \bissue  \undefined \def \bissue#1{#1}\fi
\ifx \bfpage  \undefined \def \bfpage#1{#1}\fi
\ifx \blpage  \undefined \def \blpage #1{#1}\fi
\ifx \burl  \undefined \def \burl#1{\textsf{#1}}\fi
\ifx \doiurl  \undefined \def \doiurl#1{\url{https://doi.org/#1}}\fi
\ifx \betal  \undefined \def \betal{\textit{et al.}}\fi
\ifx \binstitute  \undefined \def \binstitute#1{#1}\fi
\ifx \binstitutionaled  \undefined \def \binstitutionaled#1{#1}\fi
\ifx \bctitle  \undefined \def \bctitle#1{#1}\fi
\ifx \beditor  \undefined \def \beditor#1{#1}\fi
\ifx \bpublisher  \undefined \def \bpublisher#1{#1}\fi
\ifx \bbtitle  \undefined \def \bbtitle#1{#1}\fi
\ifx \bedition  \undefined \def \bedition#1{#1}\fi
\ifx \bseriesno  \undefined \def \bseriesno#1{#1}\fi
\ifx \blocation  \undefined \def \blocation#1{#1}\fi
\ifx \bsertitle  \undefined \def \bsertitle#1{#1}\fi
\ifx \bsnm \undefined \def \bsnm#1{#1}\fi
\ifx \bsuffix \undefined \def \bsuffix#1{#1}\fi
\ifx \bparticle \undefined \def \bparticle#1{#1}\fi
\ifx \barticle \undefined \def \barticle#1{#1}\fi
\bibcommenthead
\ifx \bconfdate \undefined \def \bconfdate #1{#1}\fi
\ifx \botherref \undefined \def \botherref #1{#1}\fi
\ifx \url \undefined \def \url#1{\textsf{#1}}\fi
\ifx \bchapter \undefined \def \bchapter#1{#1}\fi
\ifx \bbook \undefined \def \bbook#1{#1}\fi
\ifx \bcomment \undefined \def \bcomment#1{#1}\fi
\ifx \oauthor \undefined \def \oauthor#1{#1}\fi
\ifx \citeauthoryear \undefined \def \citeauthoryear#1{#1}\fi
\ifx \endbibitem  \undefined \def \endbibitem {}\fi
\ifx \bconflocation  \undefined \def \bconflocation#1{#1}\fi
\ifx \arxivurl  \undefined \def \arxivurl#1{\textsf{#1}}\fi
\csname PreBibitemsHook\endcsname

\bibitem[\protect\citeauthoryear{Myers and Myers}{1980}]{myers1980gifts}
\begin{botherref}
\oauthor{\bsnm{Myers}, \binits{I.B.}},
\oauthor{\bsnm{Myers}, \binits{P.B.}}:
{Gifts Differing: Understanding Personality Type}
(1980)
\end{botherref}
\endbibitem

\bibitem[\protect\citeauthoryear{Vaswani et~al.}{2017}]{vaswani2017attention}
\begin{bchapter}
\bauthor{\bsnm{Vaswani}, \binits{A.}},
\bauthor{\bsnm{Shazeer}, \binits{N.}},
\bauthor{\bsnm{Parmar}, \binits{N.}},
\bauthor{\bsnm{Uszkoreit}, \binits{J.}},
\bauthor{\bsnm{Jones}, \binits{L.}},
\bauthor{\bsnm{Gomez}, \binits{A.N.}},
\bauthor{\bsnm{Kaiser}, \binits{L.}},
\bauthor{\bsnm{Polosukhin}, \binits{I.}}:
\bctitle{{Attention is All You Need}}.
In: \bbtitle{Advances in Neural Information Processing Systems},
vol. \bseriesno{30},
pp. \bfpage{5998}--\blpage{6008}
(\byear{2017}).
\burl{https://papers.nips.cc/paper/7181-attention-is-all-you-need.pdf}
\end{bchapter}
\endbibitem

\bibitem[\protect\citeauthoryear{Devlin et~al.}{2018}]{devlin2018bert}
\begin{botherref}
\oauthor{\bsnm{Devlin}, \binits{J.}},
\oauthor{\bsnm{Chang}, \binits{M.-W.}},
\oauthor{\bsnm{Lee}, \binits{K.}},
\oauthor{\bsnm{Toutanova}, \binits{K.}}:
{BERT: Pre-training of Deep Bidirectional Transformers for Language Understanding}
(2018)
{\href{https://arxiv.org/abs/1810.04805}{{arXiv:1810.04805}}}
{[cs.CL]}
\end{botherref}
\endbibitem

\bibitem[\protect\citeauthoryear{Liu et~al.}{2019}]{liu2019roberta}
\begin{botherref}
\oauthor{\bsnm{Liu}, \binits{Y.}},
\oauthor{\bsnm{Ott}, \binits{M.}},
\oauthor{\bsnm{Goyal}, \binits{N.}},
\oauthor{\bsnm{Du}, \binits{J.}},
\oauthor{\bsnm{Joshi}, \binits{M.}},
\oauthor{\bsnm{Chen}, \binits{D.}},
\oauthor{\bsnm{Levy}, \binits{O.}},
\oauthor{\bsnm{Lewis}, \binits{M.}},
\oauthor{\bsnm{Zettlemoyer}, \binits{L.}},
\oauthor{\bsnm{Stoyanov}, \binits{V.}}:
{RoBERTa: A Robustly Optimized BERT Pretraining Approach}
(2019)
{\href{https://arxiv.org/abs/1907.11692}{{arXiv:1907.11692}}}
{[cs.CL]}
\end{botherref}
\endbibitem

\bibitem[\protect\citeauthoryear{Radford et~al.}{2019}]{radford2019language}
\begin{botherref}
\oauthor{\bsnm{Radford}, \binits{A.}},
\oauthor{\bsnm{Wu}, \binits{J.}},
\oauthor{\bsnm{Child}, \binits{R.}},
\oauthor{\bsnm{Luan}, \binits{D.}},
\oauthor{\bsnm{Amodei}, \binits{D.}},
\oauthor{\bsnm{Sutskever}, \binits{I.}}:
{Language Models are Unsupervised Multitask Learners}
(2019).
OpenAI Blog
\end{botherref}
\endbibitem

\bibitem[\protect\citeauthoryear{Matz et~al.}{2017}]{matz2017psychological}
\begin{barticle}
\bauthor{\bsnm{Matz}, \binits{S.C.}},
\bauthor{\bsnm{Kosinski}, \binits{M.}},
\bauthor{\bsnm{Nave}, \binits{G.}},
\bauthor{\bsnm{Stillwell}, \binits{D.J.}}:
\batitle{{Psychological Targeting as an Effective Approach to Digital Mass Persuasion}}.
\bjtitle{Proceedings of the National Academy of Sciences}
\bvolume{114}(\bissue{48}),
\bfpage{12714}--\blpage{12719}
(\byear{2017})
\doiurl{10.1073/pnas.1710966114}
\end{barticle}
\endbibitem

\bibitem[\protect\citeauthoryear{De~Choudhury et~al.}{2013}]{de2013predicting}
\begin{botherref}
\oauthor{\bsnm{De~Choudhury}, \binits{M.}},
\oauthor{\bsnm{Gamon}, \binits{M.}},
\oauthor{\bsnm{Counts}, \binits{S.}},
\oauthor{\bsnm{Horvitz}, \binits{E.}}:
{Predicting Depression via Social Media}.
Proceedings of the Seventh International AAAI Conference on Weblogs and Social Media,
128--137
(2013)
\end{botherref}
\endbibitem

\bibitem[\protect\citeauthoryear{Hinds and Joinson}{2020}]{hinds2020contextual}
\begin{barticle}
\bauthor{\bsnm{Hinds}, \binits{J.}},
\bauthor{\bsnm{Joinson}, \binits{A.}}:
\batitle{{Contextual Sensitivity and the 'Informed' User: Exploring the Impact of Privacy Controls on the Privacy Paradox}}.
\bjtitle{Journal of Broadcasting \& Electronic Media}
\bvolume{64}(\bissue{4}),
\bfpage{592}--\blpage{614}
(\byear{2020})
\doiurl{10.1080/08838151.2020.1834297}
\end{barticle}
\endbibitem

\bibitem[\protect\citeauthoryear{Ashraf and Naz}{2024}]{ashraf2024enhancing}
\begin{botherref}
\oauthor{\bsnm{Ashraf}, \binits{N.}},
\oauthor{\bsnm{Naz}, \binits{S.}}:
{Enhancing MBTI Personality Prediction from Text Data with Advanced Word Embedding Technique}.
VFAST Transactions on Software Engineering
(2024)
\end{botherref}
\endbibitem

\bibitem[\protect\citeauthoryear{Li and Che}{2024}]{li2024mbtibench}
\begin{botherref}
\oauthor{\bsnm{Li}, \binits{B.}},
\oauthor{\bsnm{Che}, \binits{W.}}:
{Can Large Language Models Understand You Better? An MBTI Personality Detection Dataset Aligned with Population Traits}.
ArXiv
(2024)
{\href{https://arxiv.org/abs/2401.12345}{{arXiv:2401.12345}}}
{[cs.CL]}
\end{botherref}
\endbibitem

\bibitem[\protect\citeauthoryear{Vásquez and Ochoa~Luna}{2021}]{vasquez2021transformer}
\begin{bchapter}
\bauthor{\bsnm{Vásquez}, \binits{R.}},
\bauthor{\bsnm{Ochoa~Luna}, \binits{J.E.}}:
\bctitle{{Transformer-based Approaches for Personality Detection using the MBTI Model}}.
In: \bbtitle{2021 XLVII Latin American Computing Conference (CLEI)}
(\byear{2021})
\end{bchapter}
\endbibitem

\bibitem[\protect\citeauthoryear{Garcia~dos Santos and Paraboni}{2022}]{dosSantos2022myers}
\begin{botherref}
\oauthor{\bsnm{Santos}, \binits{V.}},
\oauthor{\bsnm{Paraboni}, \binits{I.}}:
{Myers-Briggs personality classification from social media text using pre-trained language models}.
J. Univers. Comput. Sci.
(2022)
\end{botherref}
\endbibitem

\bibitem[\protect\citeauthoryear{Kadambi}{2021}]{kadambi2021exploring}
\begin{botherref}
\oauthor{\bsnm{Kadambi}, \binits{P.}}:
{Exploring Personality and Online Social Engagement: An Investigation of MBTI Users on Twitter}.
ArXiv
(2021)
{\href{https://arxiv.org/abs/2105.09347}{{arXiv:2105.09347}}}
{[cs.CL]}
\end{botherref}
\endbibitem

\bibitem[\protect\citeauthoryear{Bin~Tareaf}{2022}]{tareaf2022mbti}
\begin{bchapter}
\bauthor{\bsnm{Bin~Tareaf}, \binits{R.}}:
\bctitle{{MBTI BERT: A Transformer-Based Machine Learning Approach Using MBTI Model For Textual Inputs}}.
In: \bbtitle{2022 IEEE 24th International Conference on High Performance Computing \& Communications (HPCC) and Associated Conferences}
(\byear{2022})
\end{bchapter}
\endbibitem

\bibitem[\protect\citeauthoryear{Julianda and Maharani}{2023}]{julianda2023personality}
\begin{botherref}
\oauthor{\bsnm{Julianda}, \binits{A.R.}},
\oauthor{\bsnm{Maharani}, \binits{W.}}:
{Personality Detection on Reddit Using DistilBERT}.
Jurnal RESTI (Rekayasa Sistem dan Teknologi Informasi)
(2023)
\end{botherref}
\endbibitem

\bibitem[\protect\citeauthoryear{Vonitsanos and Mylonas}{2023}]{vonitsanos2023gender}
\begin{bchapter}
\bauthor{\bsnm{Vonitsanos}, \binits{G.}},
\bauthor{\bsnm{Mylonas}, \binits{P.}}:
\bctitle{{Decoding Gender on Social Networks: An In-depth Analysis of Language in Online Discussions Using Natural Language Processing and Machine Learning}}.
In: \bbtitle{2023 IEEE International Conference on Big Data (BigData)}
(\byear{2023})
\end{bchapter}
\endbibitem

\bibitem[\protect\citeauthoryear{Arya and Nishitha~D’Souza}{2023}]{arya2023prediction}
\begin{bchapter}
\bauthor{\bsnm{Arya}, \binits{S.}},
\bauthor{\bsnm{Nishitha~D’Souza}, \binits{J.}}:
\bctitle{{Prediction of MBTI with textual data using different pre-trained transformer models}}.
In: \bbtitle{2023 Fourth International Conference on Smart Technologies in Computing, Electrical and Electronics (ICSTCEE)}
(\byear{2023})
\end{bchapter}
\endbibitem

\bibitem[\protect\citeauthoryear{Harrington and Loffredo}{2010}]{harrington2010mbti}
\begin{barticle}
\bauthor{\bsnm{Harrington}, \binits{R.}},
\bauthor{\bsnm{Loffredo}, \binits{D.A.}}:
\batitle{Mbti personality type and other factors that relate to preference for online versus face-to-face instruction}.
\bjtitle{The Internet and Higher Education}
\bvolume{13}(\bissue{1-2}),
\bfpage{89}--\blpage{95}
(\byear{2010})
\end{barticle}
\endbibitem

\end{thebibliography}

\end{document}